\pdfoutput=1
\documentclass[11pt]{article}

\usepackage{emnlp2021}

\usepackage{times}
\usepackage{latexsym}

\usepackage[T1]{fontenc}
\usepackage[utf8]{inputenc}

\usepackage{microtype}

\usepackage{latexsym}
\usepackage{tabularx}
\usepackage{booktabs}
\usepackage{graphicx}
\usepackage{hyperref}
\usepackage{xcolor}
\usepackage{colortbl}
\usepackage{soul}
\usepackage{amsfonts}
\usepackage{stmaryrd}
\usepackage[T1]{fontenc}
\usepackage{enumitem}
\usepackage{hyperref}
\usepackage{afterpage}
\usepackage{amsmath}
\usepackage{amssymb}
\usepackage{multirow} 
\usepackage{balance}
\usepackage{subcaption}
\usepackage{stfloats}
\usepackage{pifont} 
\usepackage{rotating}
\usepackage{makecell}
\usepackage{hhline}
\usepackage{inconsolata}

\newcommand{\cmark}{\ding{51}}

\definecolor{darkspringgreen}{rgb}{0.09, 0.45, 0.27}
\definecolor{darkred}{rgb}{0.55, 0.0, 0.0}

\definecolor{alizarin}{rgb}{0.82, 0.1, 0.26}
\definecolor{ao(english)}{rgb}{0.0, 0.5, 0.0}
\definecolor{cadmiumgreen}{rgb}{0.0, 0.42, 0.24}

\newcommand{\kb}{KB}

\newcommand{\lm}{LM}

\newcommand{\nlp}{NLP}

\newcommand{\lama}{\textsc{LAMA}}

\newcommand{\newpar}[1]{\vspace{2mm}\noindent\textbf{#1}}

\title{Relational World Knowledge Representation \\
in Contextual Language Models: A Review} 

\author{
  Tara Safavi, Danai Koutra \\
  University of Michigan, Ann Arbor \\
  \texttt{\{tsafavi,dkoutra\}@umich.edu} \\ 
}

\begin{document}
\maketitle
\begin{abstract}
Relational \textbf{knowledge bases} (KBs) are commonly used to represent world knowledge in machines. 
However, while advantageous for their high degree of precision and interpretability, 
KBs are usually organized according to manually-defined schemas, which limit their expressiveness and require significant human efforts to engineer and maintain. 
In this review, we take a natural language processing perspective to these limitations, examining how they may be addressed in part by training deep contextual \textbf{language models} (LMs) to internalize and express relational knowledge in more flexible forms.  
We propose to organize knowledge representation strategies in LMs by the level of KB supervision provided, from no KB supervision at all to entity- and relation-level supervision. 
Our contributions are threefold: 
\textbf{(1)}~We provide a high-level, extensible taxonomy for knowledge representation in LMs;
\textbf{(2})~Within our taxonomy, we highlight notable models, evaluation tasks, and findings, in order to provide an up-to-date review of current knowledge representation capabilities in LMs; and
\textbf{(3)}~We suggest future research directions that build upon the complementary aspects of LMs and KBs as knowledge representations. 
\end{abstract}

\section{Introduction}
\label{sec:intro}
Knowledge bases (\textbf{\kb{}s}) are data structures that connect pairs of entities or concepts by semantically meaningful symbolic relations. 
Decades' worth of research have been invested into using \kb{}s as tools for relational world knowledge representation in machines~\cite{minsky-1974-framework,lenat-1995-cyc,liu-singh-2004-conceptnet,bollacker-etal-2008-freebase,vrandevcic-krotzch-2014-wikidata,speer-etal-2017-conceptnet,sap-etal-2019-atomic,ilievski-etal-2020-cskg}. 

Most large-scale modern \kb{}s are organized according to a manually engineered schema that specifies which entity and relation types are permitted, and how such types may interact with one another. 
This explicit enforcement of relational structure is both an advantage and a drawback~\cite{halevy-etal-2003-crossing}. 
On one hand, schemas support complex queries over the data with accurate, consistent, and interpretable answers. 
On the other hand, schemas are ``ontological commitments''~\cite{davis-etal-1993-what-is} that limit flexibility in how  knowledge is stored, expressed, and accessed.
Handcrafted schemas also require significant human engineering effort to construct and maintain, and are therefore often highly  incomplete~\cite{weikum-etal-2020-machine}.

\paragraph{Language models as \kb{}s?}
The tension between structured and unstructured knowledge representations is not new in natural language processing~\cite{banko-etzioni-2008-tradeoffs,fader-etal-2011-identifying}.
However, only recently has an especially promising solution emerged, brought about by breakthroughs in machine learning software, hardware, and data.
Specifically, deep contextual language models (\textbf{\lm{}s}) like BERT~\cite{devlin-etal-2019-bert} and GPT-3~\cite{brown-etal-2020-few-shot} have shown to be capable of internalizing a degree of relational world knowledge within their parameters, and expressing this knowledge across various mediums and tasks---in some cases, \emph{without} the need for a predefined entity-relation schema~\cite{petroni-etal-2019-language,roberts-etal-2020-much}. 
Consequently, some have begun to wonder whether \lm{}s will partially or even fully replace \kb{}s, given sufficiently large training budgets and parameter capacities. 

\begin{figure}[t!]
    \centering
    \includegraphics[width=0.99\columnwidth]{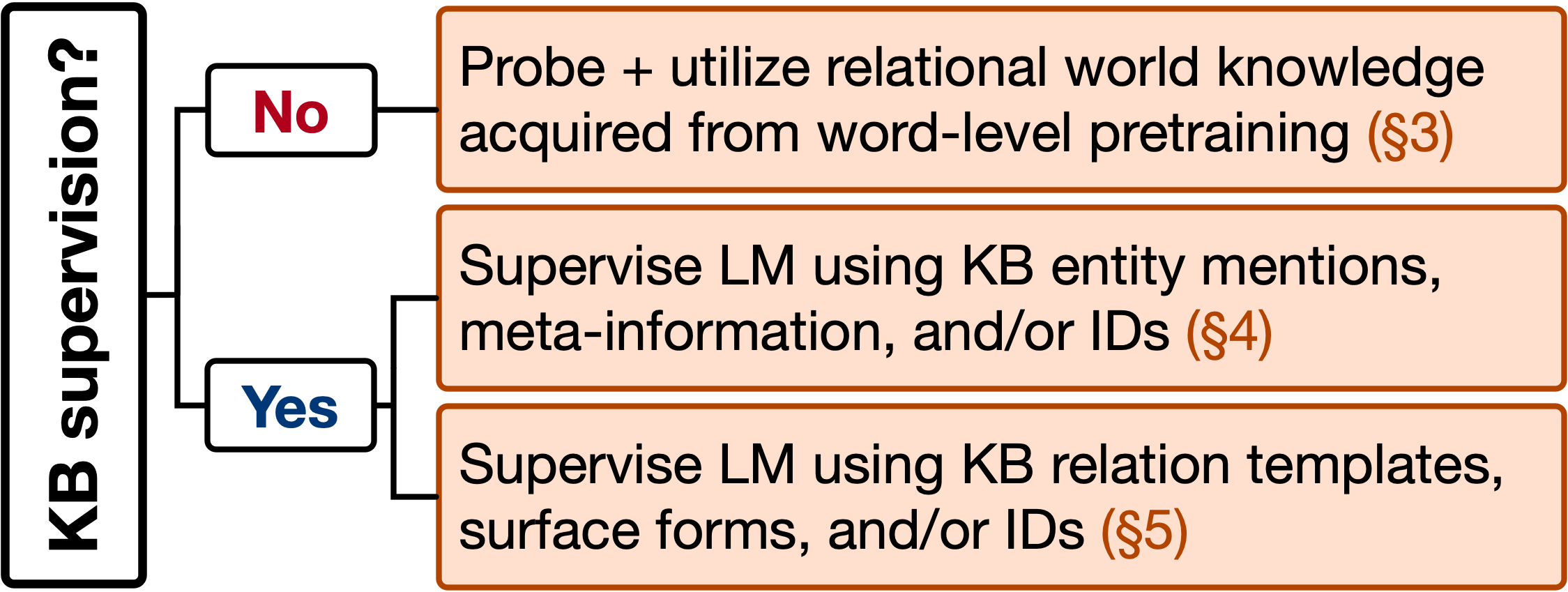}
    \caption{A high-level overview of our taxonomy, organized by level of \kb{} supervision provided. 
    }
    \label{fig:taxonomy-high-level}
    \vspace{-.3cm}
\end{figure}

\paragraph{Present work}
In this review, we summarize recent compelling progress in machine representation of relational world knowledge with \lm{}s.
We propose to organize relevant work by the level of \kb{} supervision provided to the \lm{} (Figure~\ref{fig:taxonomy-high-level}): 
\begin{itemize}
    \setlength{\itemsep}{0.001em}
    \item \textbf{Word-level supervision} (\S~\ref{sec:implicit}): 
    At this level, \lm{}s are not explicitly supervised on a \kb, but may be indirectly exposed to \kb-like knowledge via word associations in the training corpus. 
    Here, we cover techniques for probing and utilizing  this implicitly acquired knowledge.
    \item \textbf{Entity-level supervision} (\S~\ref{sec:entities}): 
    At this level, \lm{}s are supervised to acquire knowledge of \kb{} entities. 
    Here, we organize strategies from ``less symbolic'' to ``more symbolic'': 
    Less symbolic approaches train \lm{}s with entity-aware language modeling losses, but never explicitly require the \lm{} to link entity mentions to the \kb. 
    By contrast, more symbolic approaches involve linking, and may also integrate entity embeddings into the \lm{}'s parameters. 
    \item \textbf{Relation-level supervision} (\S~\ref{sec:relations}):
    At this level, \lm{}s are supervised to acquire knowledge of \kb{} triples and paths. 
    Again, we organize strategies from less to more symbolic, where less symbolic approaches treat triples as fully natural language statements, and more symbolic approaches incorporate dedicated embeddings of \kb{} relation types.  
\end{itemize}
For each supervision level, we provide notable examples in terms of methodology and/or findings, and compare the benefits and drawbacks of different approaches.
We conclude in \S~\ref{sec:conclusion} with our vision of the future, emphasizing the complementary roles of \lm{}s and \kb{}s as knowledge representations. 

\paragraph{Related work}
As this topic is relatively nascent, few related surveys exist. 
Closest to our own work, 
\citet{colon-etal-2021-combining} cover methods for combining contextual language representations with graph representations, albeit with a comparatively narrow scope and no discussion of implicit knowledge. %  or probing.
\citet{liu-etal-2021-prompting-survey} survey prompt-based learning in \lm{}s, which overlaps with our discussion of cloze prompting  in \S~\ref{implicit:cloze}, although relational world knowledge is not their main focus.

\section{Preliminaries}
\label{sec:prelim}
% \begin{table*}[t!]
%     \centering
%     \caption{A comparison of contextual \lm{}s and \kb{}s as world knowledge representations. }
%     \label{table:kglm-comparison}
%     \resizebox{\textwidth}{!}{
%     \begin{tabular}{l p{7.3cm} p{7.3cm} }
%         \toprule
%         & Knowledge base (\kb) & Contextual language model (\lm) \\ 
%         \toprule
%         Organization &  Graph-like data structure that connects pairs of symbolic entities via symbolic relations according to a predefined schema & Matrices of parameter weights learned (optimized) via pretraining and possibly fine-tuning, organized according to a neural architecture  \\
%         \midrule 
%         % \multirow{2}{*}{\rot{90}{Acquisition} 
%         Knowledge acquisition & Explicit population via manual curation and/or automatic information extraction & 
%         Implicit (self-supervised pretraining); possibly explicit (supervised fine-tuning) \\ 
%         Knowledge accuracy & Human-curated \kb{}s (Wikidata, ConceptNet) are relatively high-precision but low-recall &
%         Currently, precision is relatively low, and recall scales linearly with parameter capacity \\ 
%         \midrule 
%         Query input & Structured, disambiguated queries potentially mapped from natural language & Arbitrary non-disambiguated sequences, e.g., natural language prefixes, prompts, or questions  \\ 
%         % \midrule 
%         Query output & Symbolic entities, relations, and/or structures & Tokens, sequences, and/or labels \\ 
%         \bottomrule
%     \end{tabular}
%     }
% \end{table*}

We briefly review preliminaries and assumptions necessary for our survey. %  of the literature.

\paragraph{Knowledge bases}
We use the term ``knowledge base'' (\textbf{\kb}) to refer to a relational data structure comprising a set of \textbf{entities} $E$, \textbf{relation types} $R$, and \textbf{triples} $(s, r, o) \in E \times R \times E$, where $s, o \in E$ are subject and object entities, respectively.\footnote{For our purposes, we consider the terms ``knowledge base'' and ``knowledge graph'' as interchangeable.} %  and $r \in R$ is a pairwise relation.
We consider two types of \kb{}s under the umbrella of ``relational world knowledge.'' 
\textbf{Encyclopedic} \kb{}s store facts about typed, disambiguated entities; a well-known example is the Wikidata \kb~\cite{vrandevcic-krotzch-2014-wikidata}, which, like its sister project Wikipedia, is publicly accessible and collaboratively constructed. 
By contrast, in \textbf{commonsense} \kb{}s, ``entities'' are typically represented by non-canonicalized free-text phrases.
% phrases referring to, e.g., everyday objects and actions. 
Examples include the publicly accessible, crowdsourced {ConceptNet}~\cite{liu-singh-2004-conceptnet,speer-etal-2017-conceptnet} and ATOMIC~\cite{sap-etal-2019-atomic} \kb{}s.
% \footnote{``Commonsense''  is loosely defined and may overlap with encyclopedic knowledge in some domains, for example geographical knowledge~\cite{lobue-yates-2011-types}.}
% We refer the reader to~\cite{weikum-etal-2020-machine} for more details on \kb{} construction. 

% \begin{figure}[t!]
%     \centering
%     \includegraphics[width=0.75\columnwidth]{fig/triple-examples.png}
%     \caption{Illustrative examples of knowledge triples from encyclopedic and commonsense \kb{}s. 
%     }
%     \label{fig:triples}
%     \vspace{-.3cm}
% \end{figure}

% The two major goals of \kb{}s, correctness (precision) and coverage (recall), are in constant tension~\cite{weikum-etal-2020-machine}. 
% The primary \kb{} resources that we consider (Wikidata, ConceptNet, ATOMIC) are relatively high-precision and are used as gold standards or supervision sources for many applications in natural language processing. 
% However, they tend to be low-recall for all but the most popular of entities, albeit continually expanding~\cite{razniewski-etal-2020-structured}. 

\paragraph{Language models}
Following the contemporary \nlp{} literature, we use the term ``language model'' (\textbf{\lm{}}) to refer to a deep neural network that is trained to learn contextual text representations. %  that is, each token's representation depends on its surrounding textual context~\cite{peters-etal-2018-deep}.
\lm{}s generally come \textbf{pretrained}, with parameters pre-initialized for generic text representation via self-supervised training on large corpora, and may be used as-is after pretraining,  or further \textbf{fine-tuned} with supervision on downstream task(s). 
This work considers \lm{}s based on the \textbf{Transformer} architecture~\cite{vaswani-etal-2017-attention}, examples of which include the 
encoder-only BERT family~\cite{devlin-etal-2019-bert,liu-etal-2019-roberta}, 
the decoder-only GPT family~\cite{brown-etal-2020-few-shot}, and
the encoder-decoder T5~\cite{raffel-etal-2020-exploring} and BART~\cite{lewis-etal-2020-bart} families. 
% We refer the reader to~\cite{qiu-etal-2020-survey} for more details on pretrained \lm{}s, and~\cite{rush-2018-annotated} for more details on the Transformer architecture. 

\section{Word-level supervision}
\label{sec:implicit}
The standard language modeling task is to predict the $n$-th word in a sequence of $n$ words---that is, a conditional probability estimation task~\cite{radford-etal-2019-language}. 
While many variants of this task have been proposed to allow \lm{}s to condition their predictions on different inputs~\cite{devlin-etal-2019-bert,raffel-etal-2020-exploring,lewis-etal-2020-bart}, a notable feature of all such approaches is that they operate at the word (and subword) level.

If these supervision techniques do not incorporate \kb{}s at all, how are they relevant when considering \lm{}s as relational knowledge representations?
The answer is simple.  
Typical language modeling corpora like Wikipedia are known to contain \kb-like assertions about the world~\cite{da-kasai-2019-cracking}.
\lm{}s trained on enough such data can be expected to acquire some \kb-like knowledge, even without targeted entity- or relation-level supervision. 
Therefore, in order to motivate the necessity (if at all) of \kb{} supervision, it is crucial to first understand what relational world ``knowledge'' \lm{}s acquire from word-level pretraining. %  alone.
In this section, we cover strategies to extract and utilize this knowledge under the cloze prompting (\S~\ref{implicit:cloze}) and statement scoring (\S~\ref{implicit:scoring}) protocols.  Table~\ref{table:implicit} provides a taxonomy for this section, with representative examples and evaluation tasks.

\begin{table*}[t!]
\centering
\caption{
    Taxonomy and representative examples for extracting relational knowledge in word-level pretrained \lm{}s, with evaluation tasks that have been conducted in the referenced papers. 
    \emph{Glossary of evaluation tasks}: KP---knowledge probing; QA---question answering; CR---compositional reasoning; KC---knowledge base construction.
}
\label{table:implicit}
\resizebox{0.95\textwidth}{!}{
    \begin{tabular}{ ll p{7cm} c cccc }
    \toprule
    \multirow{2}{*}{\textbf{Knowledge extracted via...}} & \multirow{2}{*}{\textbf{Extraction strategy}} & \multirow{2}{*}{\textbf{Representative examples}} && \multicolumn{4}{c}{\textbf{Evaluation task(s)}} \\ 
    \cline{5-8}
    & & && KP & QA & CR & KC \\ 
    \toprule 
    \multirow{9}{*}{Cloze prompts (\S~\ref{implicit:cloze})}
        & \multirow{1}{*}{Prompt handcrafting} & \cite{petroni-etal-2019-language, dufter-etal-2021-static}  && \multirow{1}{*}{\cmark} & & & \\ 
        \cline{2-8}
        & \multirow{2}{*}{Automatic prompt engineering} 
        & \cite{jiang-etal-2020-how-can,shin-etal-2020-autoprompt,zhong-etal-2021-factual,qin-eisner-2021-learning} && \multirow{2}{*}{\cmark} & & & \\ 
        \cline{2-8}
        & \multirow{2}{*}{Adversarial prompt modification} 
        & \cite{kassner-schutze-2020-negated,petroni-etal-2020-context,poerner-etal-2020-e-bert,cao-etal-2021-knowledgeable} && \multirow{2}{*}{\cmark} & & & \\ 
        \cline{2-8}
        & \multirow{2}{*}{Varying base prompts} & \cite{elazar-2021-measuring-improving,heinzerling-inui-2021-language,jiang-etal-2020-x-factr,kassner-etal-2021-multilingual} && \multirow{2}{*}{\cmark} & & & \\ 
        \cline{2-8}
        & Symbolic rule-based prompting &  \cite{kassner-etal-2020-pretrained,talmor-etal-2020-olmpics} && \cmark & & \cmark & \\ 
    \midrule 
    \multirow{2}{*}{Statement scores (\S~\ref{implicit:scoring})}
        & Single-\lm{} scoring & \cite{tamborrino-etal-2020-pre,zhou-etal-2020-evaluating} && & \cmark & \cmark & \\ 
        \cline{2-8}
        & Dual-\lm{} scoring & \cite{davison-etal-2019-commonsense,shwartz-etal-2020-unsupervised} && & \cmark & & \cmark \\ 
    \bottomrule
\end{tabular}
}
\vspace{-.3cm}
\end{table*}

\subsection{Cloze prompting}
\label{implicit:cloze}

\begin{figure}[t!]
    \centering
    \includegraphics[width=0.91\columnwidth]{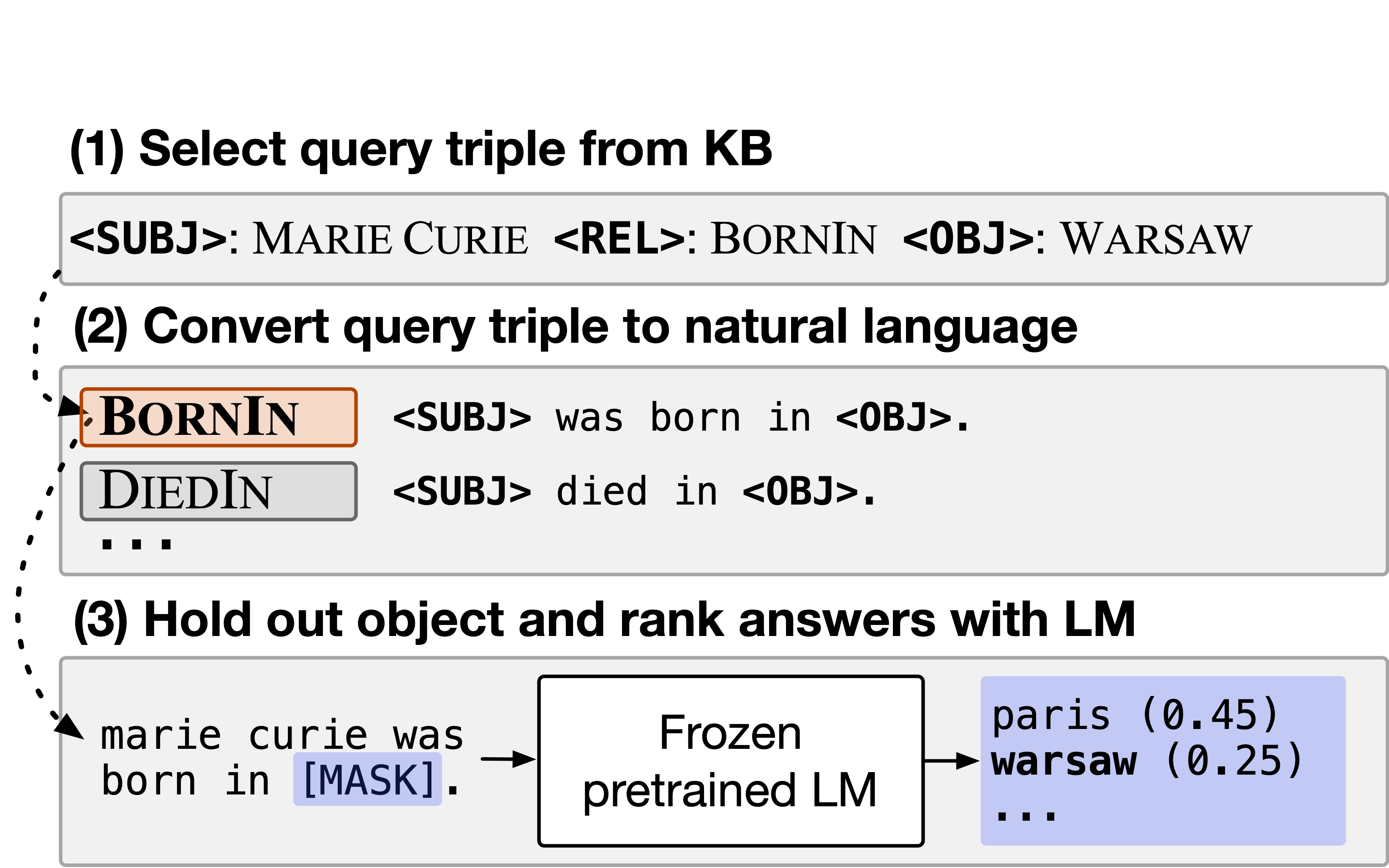}
    \caption{Probing relational knowledge in pretrained \lm{}s with cloze prompts generated from \kb{} triples. 
    }
    \label{fig:cloze}
    \vspace{-.5cm}
\end{figure}

The cloze prompting protocol
(\citealp{taylor-1953-cloze} and Figure~\ref{fig:cloze}) is a direct approach for extracting and evaluating \kb-like knowledge in pretrained \lm{}s.
Under this protocol, 
\kb{} triples are first converted to natural language assertions using (e.g.) relation templates.  
For each assertion, the token(s) corresponding to the object entity are held out. %  a few relevant relational statements or a text passage may also be prepended to the prompt to ``prime'' the \lm~\cite{brown-etal-2020-few-shot}. 
A frozen pretrained \lm{} then ranks candidate tokens within its vocabulary by the probability that they fill in the empty slot(s). 
Accuracy is typically measured by the proportion of prompts for which the correct answer appears in the \lm{}'s top-$k$ predictions, with the assumption that better performance implies more pretrained knowledge within the \lm{}. 
%\footnote{This metric has been called  ``precision@$k$'', ``hits@$k$'', or simply ``top-$k$.''  Note, however, that precision@$k$ has a different meaning in information retrieval~\cite{schutze-etal-2008-introduction}.}
% \footnote{This protocol has an entity-level analogue in link prediction (LP): 
% In LP, a  set of \kb{} triples missing their subject or object entities are presented to a model, which ranks \kb{} entities in order of likelihoood that they complete the triple.}
% An LP model must then rank \kb{} entities in order of predicted likelihood that they complete the triple~\cite{bordes-etal-2013-translating}.}

\newpar{Handcrafted prompts} in English with single-token answers make up \lama{}~\cite{petroni-etal-2019-language}, one of the earliest and most widely-used \lm{} cloze probes. 
\lama{}, which is mapped primarily to Wikidata and ConceptNet triples, was initially used to compare pretrained \lm{}s' knowledge  to off-the-shelf \kb{} question answering systems. 
\citet{petroni-etal-2019-language} showed that pretrained BERT is competitive with a supervised relation extraction model that has been provided an oracle for entity linking,  particularly for 1-1 queries.
Subsequent work has experimented with handcrafted templates for probing the knowledge of both very large (hundred-billion parameter) \lm{}s~\cite{brown-etal-2020-few-shot} as well as non-contextual word embeddings, i.e., as a simple control baseline for \lm{}s~\cite{dufter-etal-2021-static}.
Both studies demonstrate some success, particularly in cases where the probed model is provided a small amount of extra context in the form of conditioning examples~\cite{brown-etal-2020-few-shot} or entity type information~\cite{dufter-etal-2021-static}.

\newpar{Automatic prompt engineering} is a promising alternative to prompt handcrafting for knowledge extraction in \lm{}s~\cite{liu-etal-2021-prompting-survey}, as prompts engineered using discrete~\cite{jiang-etal-2020-how-can,shin-etal-2020-autoprompt,haviv-etal-2021-bertese} and continuous~\cite{zhong-etal-2021-factual,qin-eisner-2021-learning,liu-etal-2021-gpt} optimization have improved \lm{}s' lower-bound performance on \lama{}'s underlying queries. %  in zero-shot and few-shot settings~\cite{brown-etal-2020-few-shot}. 
Note, however, that optimized prompts are not always grammatical or intelligible~\cite{shin-etal-2020-autoprompt}. 
Prompt optimization methods may also confound knowledge probes by overfitting to the probes' answer distributions during training~\cite{zhong-etal-2021-factual,cao-etal-2021-knowledgeable}, and often require large validation sets for tuning, which may not be feasible in practice~\cite{perez-etal-2021-true}.

\newpar{Adversarial modification} of \lama{} prompts has uncovered weaknesses in pretrained \lm{}s' world ``knowledge,'' for example that BERT's accuracy drops precipitously when irrelevant statements or negation words are added to prompts~\cite{kassner-schutze-2020-negated,lin-etal-2020-birds,petroni-etal-2020-context}, and that it can ``guess'' answers using shallow lexical cues or benchmark artifacts~\cite{poerner-etal-2020-e-bert,cao-etal-2021-knowledgeable}. 
However, the adversarial robustness of \lm{} knowledge improves greatly with supervision in both the pretraining~\cite{petroni-etal-2020-context} and fine-tuning~\cite{kassner-schutze-2020-negated} stages, suggesting that explicit \kb{}-level supervision is a viable remedy to input sensitivity. 

\newpar{Several collections of prompt variations}, including paraphrased sets of base prompts~\cite{elazar-2021-measuring-improving,heinzerling-inui-2021-language} and multilingual sets of base (English) prompts~\cite{jiang-etal-2020-x-factr,kassner-etal-2021-multilingual} have been released to expand the original research questions posed by \lama.
For the former, it has been found that pretrained BERT-based \lm{}s typically do not output consistent answers for prompt paraphrases, although their consistency can again be greatly improved by targeted pretraining~\cite{elazar-2021-measuring-improving,heinzerling-inui-2021-language}.
For the latter, initial results on prompts beyond English indicate high variability in pretrained \lm{} performance across languages and poor performance on prompts with multi-token answers~\cite{jiang-etal-2020-x-factr,kassner-etal-2021-multilingual}.  

\newpar{Prompts generated with symbolic rules} have been used to test pretrained \lm{}s' abilities to learn, e.g., equivalence, implication, composition, and conjunction.
Existing studies vary the degrees of experimental control: 
\citet{talmor-etal-2020-olmpics} use BERT-based models with their publicly-available pretrained weights, whereas \citet{kassner-etal-2020-pretrained} pretrain BERT from scratch on synthetic \kb{} triples only.
Both studies observe mixed results, concluding that word-level pretraining alone (at least on BERT) does not lead to strong ``reasoning'' skills. 
% finding that pretrained \lm{}s are capable of learnin
% and  
% When pretrained \emph{only} on synthetic \kb{} triples exhibiting symbolic rules and tested on prompts inferrable using those rules, 
% BERT appears to learn equivalence and implication, but not composition and negation~\cite{kassner-etal-2020-pretrained}. 
% Under a less-controlled setting in which BERT-based models with publicly-available pretrained weights answer cloze prompts testing comparison and conjunction, results are again mixed~\cite{talmor-etal-2020-olmpics}. 

\subsection{Statement scoring}
\label{implicit:scoring}

Beyond probing, pretrained \lm{} ``knowledge'' can be purposed toward downstream \kb-level tasks in a zero-shot manner via  statement scoring. 
Here, a pretrained \lm{} is fed natural language statements corresponding to \kb{} triples, and its token probabilities across each statement are pooled to yield statement scores.
These scores are then treated as input to a downstream decision, mirroring the way that supervised \lm{}s can be trained to output probabilities for triple-level prediction tasks (\S~\ref{sec:relations}). 
We categorize statement scoring strategies as single- or dual-\lm{} approaches.
The \textbf{single-\lm{}} approach pools the pretrained \lm{}'s token scores over a candidate set of sequences, then takes the highest-scoring sequence as the \lm{}'s ``prediction'' or choice~\cite{tamborrino-etal-2020-pre,bouraoui-etal-2020-inducing,zhou-etal-2020-evaluating,brown-etal-2020-few-shot}. 
The \textbf{dual-\lm{}} framework first uses one pretrained \lm{} to generate useful context (e.g., clarification text) for the task, then feeds this context to another, possibly different pretrained \lm{} to obtain a final score~\cite{davison-etal-2019-commonsense,shwartz-etal-2020-unsupervised}. 

Both categories have shown promise over comparable unsupervised (and, under some conditions, supervised) methods for tasks like multiple-choice QA~\cite{tamborrino-etal-2020-pre,shwartz-etal-2020-unsupervised,brown-etal-2020-few-shot} and commonsense \kb{} completion~\cite{davison-etal-2019-commonsense}. 
However, \lm{} scores have also shown to be sensitive to small perturbations in text~\cite{zhou-etal-2020-evaluating}, so this approach may be less effective on noisy or long-tail inputs. % ---again motivating the need for targeted \kb-level supervision, which has shown to improve adversarial robustness in \lm{}s in knowledge probing (\S~\ref{implicit:cloze}).  

\subsection{Summary and outlook}
\label{implicit:summary}

There is still broad disagreement over the nature of acquired ``knowledge'' in pretrained \lm{}s.
Whereas some studies suggest that word-level pretraining may be enough to endow \lm{}s with \kb-like knowledge~\cite{petroni-etal-2019-language,tamborrino-etal-2020-pre}, in particular given enough parameters and the right set of prompts~\cite{brown-etal-2020-few-shot}, others conclude that such pretraining alone does not yield sufficiently precise or robust \lm{} knowledge~\cite{elazar-2021-measuring-improving,cao-etal-2021-knowledgeable}---directly motivating the targeted supervision strategies discussed in the remainder of this paper. 
We observe that different studies independently set objectives for what a pretrained \lm{} should ``know,'' and thus naturally reach different conclusions.
We believe that future studies must reach consensus on standardized tasks and benchmarks, addressing questions like: 
What degree of overlap between a pretraining corpus and a knowledge probe is permissible, and how can this be accurately uncovered and quantified? 
What lexical cues or correlations should be allowed in knowledge probes? 
Progress in this direction will not only further our understanding of the effects of word-level supervision on \lm{} knowledge acquisition, but will also provide appropriate yardsticks for measuring the benefits of targeted entity- and relation-level supervision.

\section{Entity-level supervision}
\label{sec:entities}
% Most works discussed in the previous section indicate that while contextual \lm{}s do acquire some relational world knowledge during pretraining, such knowledge is not particularly precise or robust without targeted supervision.\footnote{This is perhaps to be expected, since ``vanilla'' \lm{}s are pretrained primarily to model token-level co-occurrences, not to store high-precision factual knowledge.}

We next review entity-level supervision strategies for \lm{}s, most often toward improving performance in knowledge probes like \lama{} (\S~\ref{implicit:cloze}) and canonical \nlp{} tasks like entity typing, entity linking, and question answering.
We roughly categorize approaches from ``least symbolic'' to ``most symbolic.'' 
On the former end of the spectrum,
the \lm{} is exposed to entity mentions in text but not required to link these mentions to an external entity bank (\S~\ref{entities:token-prediction}).
On the latter end, the \lm{} is trained to link mentions to the \kb{} using late (\S~\ref{entities:entity-linking}) or mid-to-early fusion approaches (\S~\ref{entities:embeddings}). 
Table~\ref{table:entity} provides a taxonomy of supervision strategies for this section with representative examples. 

% \begin{figure}[t!]
%     \centering
%     \includegraphics[width=0.99\columnwidth]{fig/entity-taxonomy.png}
%     \caption{A taxonomy of tasks and methods for entity-level supervision in \lm{}s (\S~\ref{sec:entities}), with representative examples per category. 
%     }
%     \label{fig:taxonomy-entities}
%     \vspace{-.5cm}
% \end{figure}
\begin{table*}[t!]
\centering
\caption{
    Taxonomy and representative examples of entity-level supervision in \lm{}s, with evaluation tasks that have been conducted in the referenced papers. 
    \emph{Glossary of evaluation tasks}: KP---knowledge probing; EL---entity linking; ET---entity typing; RC---relation classification; QA---question answering; GL---the General Language Understanding Evaluation or GLUE benchmark~\cite{wang-etal-2019-glue}, which covers multiple subtasks.
}
\label{table:entity}
\resizebox{\textwidth}{!}{
    \begin{tabular}{ l l l c cccccc}
    \toprule
    \multirow{2}{*}{\textbf{Entities as...}} & \multirow{2}{*}{\textbf{Supervision strategy}} & \multirow{2}{*}{\textbf{Representative examples}} && \multicolumn{6}{c}{\textbf{Evaluation task(s)}} \\ 
    \cline{5-10}
     & & && KP & EL & ET & RC & QA & GL   \\ 
     \toprule 
     \multirow{2}{*}{Token mention-spans (\S~\ref{entities:token-prediction})} 
        & Masked token prediction & \cite{roberts-etal-2020-much,guu-etal-2020-retrieval}
        && & & & & \cmark & \\
        & Contrastive learning & \cite{xiong-etal-2020-pretrained-encyclopedia,shen-etal-2020-exploiting} && \cmark & & \cmark & & \cmark & \\ 
     \midrule 
     \multirow{3}{*}{Text-to-\kb{} links---late fusion (\S~\ref{entities:entity-linking})} 
        & Linking w/o external info &  \cite{broscheit-2019-investigating,ling-etal-2020-learning-cross-context} && & \cmark & & & & \cmark  \\ 
        & Linking w/ textual metadata & \cite{wu-etal-2020-scalable,de-cao-2021-autoregressive} && & \cmark &  & \cmark & \cmark & \\ 
        & Linking w/ external embeddings & \cite{zhang-etal-2019-ernie,chen-etal-2020-contextualized} && & \cmark & \cmark & \cmark & & \cmark \\ 
    \midrule 
    \multirow{2}{*}{Text-to-\kb{} links---mid/early fusion (\S~\ref{entities:embeddings})} 
        & Entity embedding retrieval & \cite{peters-etal-2019-knowledge,fevry-etal-2020-entities} && \cmark & \cmark & \cmark & \cmark & \cmark & \\ 
        & Treating entities as tokens & \cite{yamada-etal-2020-luke,poerner-etal-2020-e-bert} && \cmark & \cmark & \cmark & \cmark & \cmark & \\ 
    \bottomrule
\end{tabular}
}
\vspace{-.3cm}
\end{table*}

\subsection{Modeling entities without linking}
\label{entities:token-prediction}

\begin{figure*}[t!]
    \centering
    \begin{subfigure}{.23\textwidth}
      \centering
      % include first image
      \includegraphics[width=\linewidth]{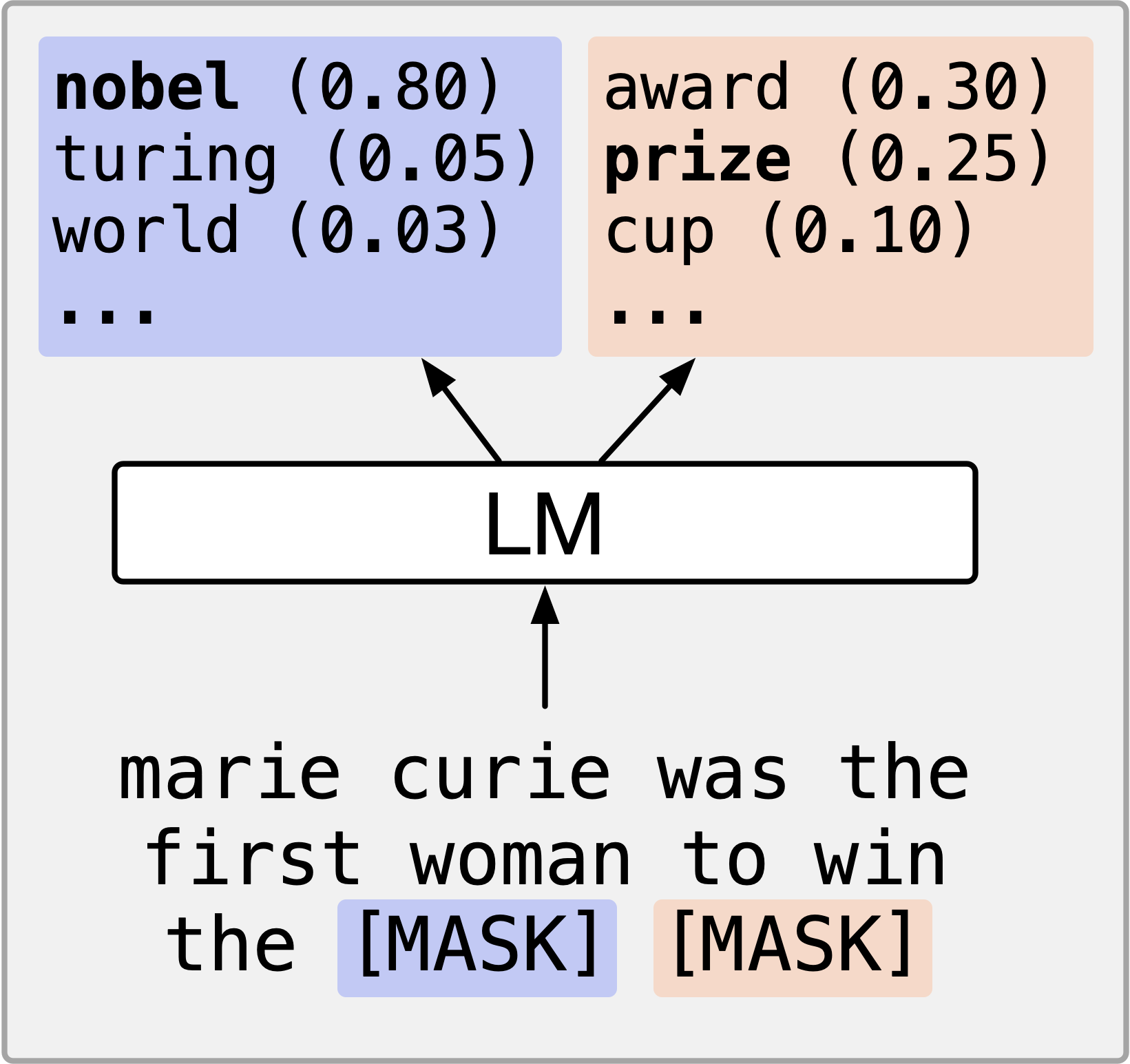}
      \caption{Mention-span masking}
      \label{fig:entity-token-prediction}
    \end{subfigure}
    ~
    \begin{subfigure}{.23\textwidth}
      \centering
      % include first image
      \includegraphics[width=\linewidth]{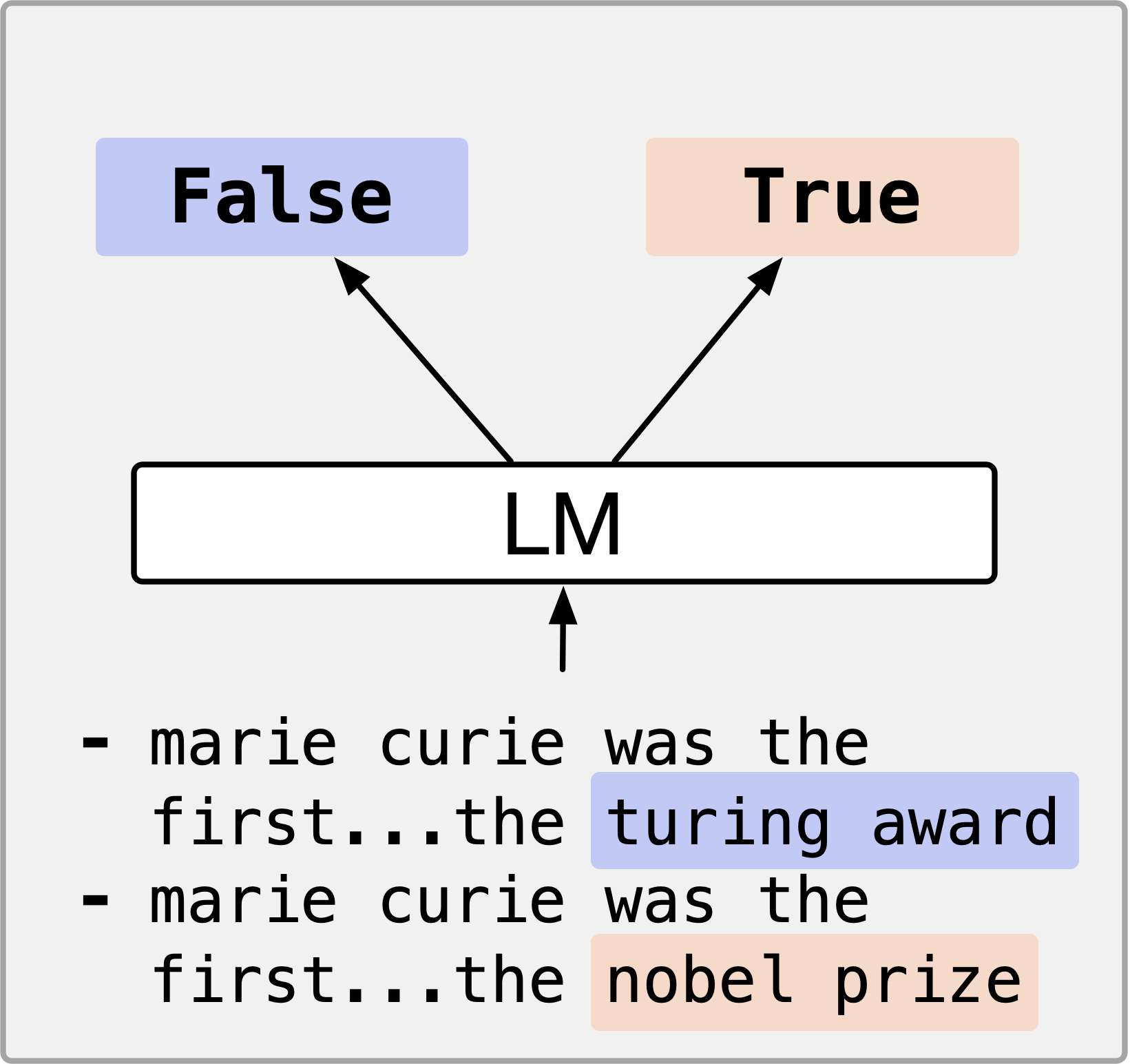}
      \caption{Contrastive learning}
      \label{fig:entities-contrastive}
    \end{subfigure}
    ~
    \begin{subfigure}{.23\textwidth}
      \centering
      % include first image
      \includegraphics[width=\linewidth]{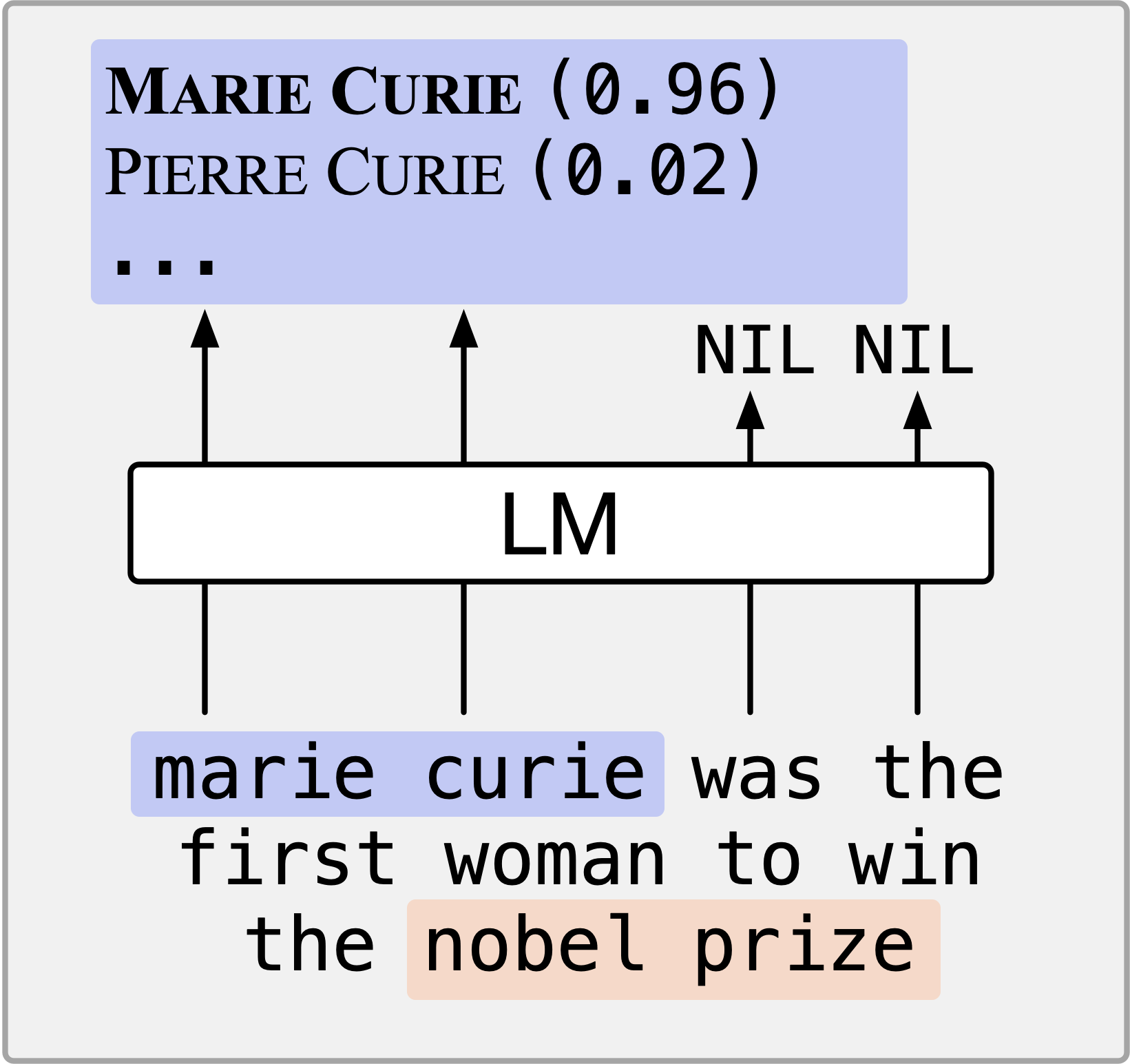}
      \caption{Linking---late fusion}
      \label{fig:entity-linking}
    \end{subfigure}
    ~
    \begin{subfigure}{.23\textwidth}
      \centering
      % include first image
      \includegraphics[width=\linewidth]{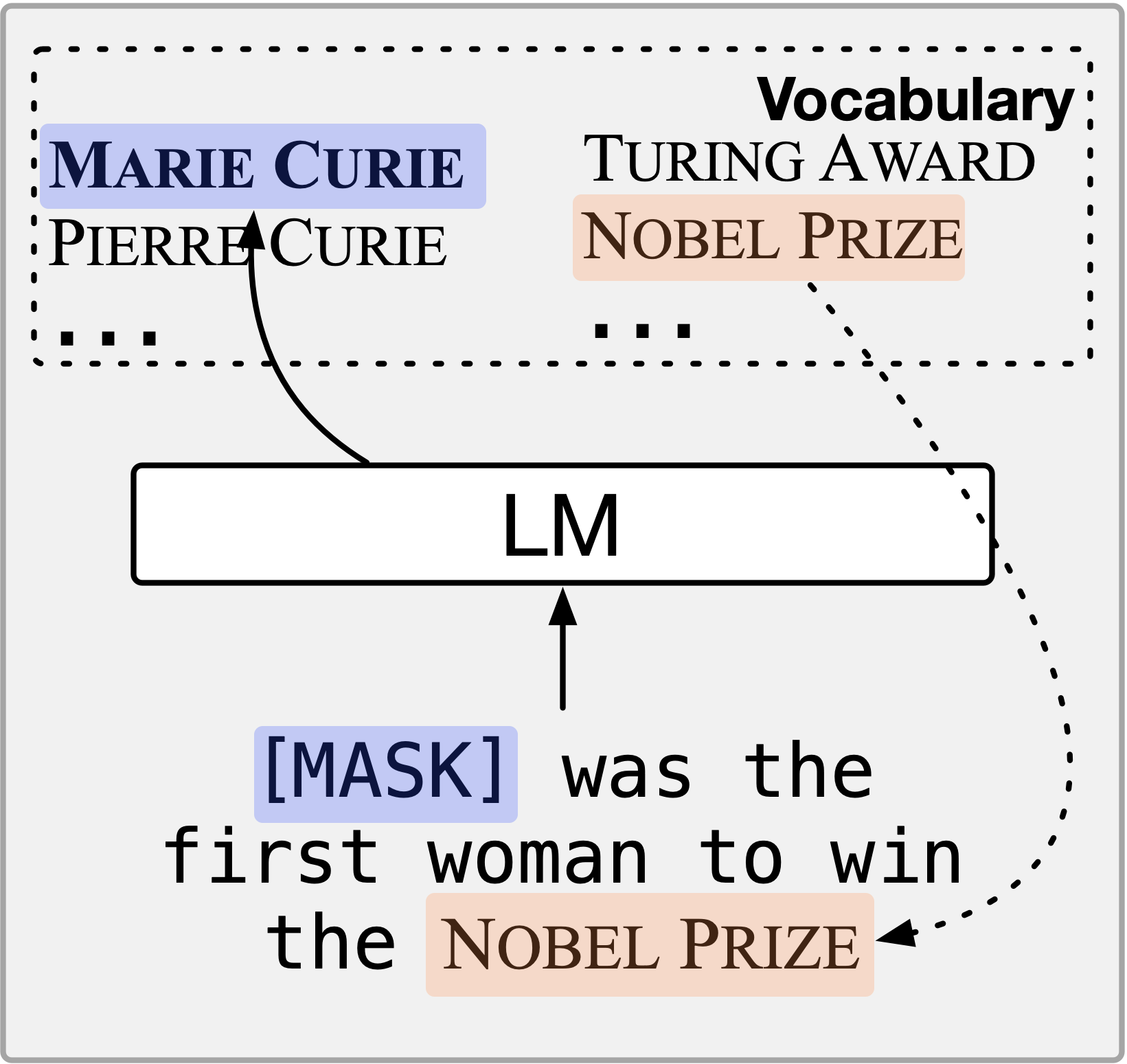}
      \caption{Linking---early fusion}
      \label{fig:entity-embeddings}
    \end{subfigure}
    \caption{Examples of entity-level supervision in \lm{}s, ranging from ``less symbolic''  to ``more symbolic.''
    }
    \label{fig:entity-supervision}
    \vspace{-.5cm}
\end{figure*}

The ``least symbolic'' entity supervision approaches that we consider input textual contexts containing entity mention-spans to the \lm{}, and incorporate these mention-spans into their losses.
However, they do not require the \lm{} to link these mentions to the \kb{}'s entity set, so the \lm{} is never directly exposed to the \kb.
Figures~\ref{fig:entity-token-prediction} and~\ref{fig:entities-contrastive} provide examples of input and output for this class of approaches.

% The first category of approaches we consider train an \lm{} to predict the tokens in mention-spans referring to known (in-\kb) entities, \NEW{without requiring the \lm{} to explicitly link these mentions to the \kb's entity set}.
% Here, all predictions are made at the token level, so the \lm{} is never directly exposed to the \kb's entity set.
% \footnote{However,  if the chosen training text is not already entity-linked, the practitioner must preprocess the corpus using a \kb{} in order to detect and link entity mentions.}

% The first class of entity supervision approaches that we cover do not directly expose the \lm{} to the \kb's entity set. 
% Rather, they train the \lm{} to model the tokens in entity \emph{mention-spans}, or contiguous sequences of tokens referring to known entities, by preprocessing the training corpus 
% into one of the \lm{}'s pretraining objectives, and require all predictions to be made at the token or token-span level. 

% \lm{}s may be pretrained on entity \emph{mention-spans}, or contiguous sequences of tokens referring to known entities. 
% In these approaches, the \lm{}'s predictions are all made at the token or token-span level, without directly exposing the \lm{} to the \kb{}'s entity set.\footnote{Note, however, that if the chosen training text is not already entity-linked, the practitioner must preprocess the corpus using a \kb{} in order to detect and link entity mentions.}

\newpar{Masking tokens in mention-spans} 
and training \lm{}s to predict these tokens may promote knowledge memorization~\cite{sun-etal-2020-ernie}.
% This hypothesis is empirically supported by
\citet{roberts-etal-2020-much} investigate this strategy using a simple masking strategy whereby an \lm{} is trained to predict the tokens comprising named entities and dates in text (Figure~\ref{fig:entity-token-prediction}, originally proposed by \citealp{guu-etal-2020-retrieval}). 
The authors find that the largest (11 billion parameter) version of T5 generates exact-match answers on open-domain question answering (QA) benchmarks with higher accuracy than extractive systems---even without access to external context documents, simulating a ``closed-book'' exam. 

\newpar{Contrastive learning} techniques, which have been used for \lm{} supervision at the word and sentence level~\cite{devlin-etal-2019-bert}, have also been devised for supervision on entity mentions~\cite{shen-etal-2020-exploiting}. 
For example, \citet{xiong-etal-2020-pretrained-encyclopedia} replace a proportion of entity mentions in the pretraining corpus with the names of negatively-sampled entities of the same type, and train an \lm{} to predict whether the entity in the span has been replaced (Figure~\ref{fig:entities-contrastive}).
Although the previously discussed closed-book T5 model~\cite{roberts-etal-2020-much} outperforms~\citet{xiong-etal-2020-pretrained-encyclopedia}'s open-book BERT pretrained with contrastive entity replacement on open-domain QA, 
the latter may generalize better: 
T5's performance degrades considerably for facts not  observed during training, whereas open-book approaches appear more robust~\cite{lewis-etal-2021-question}.

\subsection{Linking with late fusion}
\label{entities:entity-linking}

The next-strongest level of entity supervision is to train the \lm{} to link entity-centric textual contexts to a \kb{}'s entity set $E$. 
Here, we cover late fusion approaches, which operate at the word level in terms of input to the \lm{} and incorporate entities at the \lm{}'s output layer only, as exemplified in Figure~\ref{fig:entity-linking}.
The simplest representatives of this category train \lm{}s to match individual tokens~\cite{broscheit-2019-investigating} or mentions~\cite{ling-etal-2020-learning-cross-context} in a text corpus to an entity bank, without any external resources. %  or supervision signals. 
The minimally ``entity-aware'' BERT proposed by~\citet{broscheit-2019-investigating}, which adds a single classification layer on top of a pretrained BERT encoder, achieves competitive results with a state-of-the-art specialized entity linking architecture~\cite{kolitsas-etal-2018-end}.
% However, it does not improve over plain BERT in the popular General Language Understanding Evaluation (GLUE) benchmark~\cite{wang-etal-2019-glue}, echoing concurrent and subsequent work that finds that GLUE does not require world knowledge~\cite{zhang-etal-2019-ernie,wang-etal-2020-kepler,lauscher-etal-2020-common}.

\newpar{Entity meta-information} such as names and descriptions are viable external resources for \lm-powered entity linking~\cite{botha-etal-2020-entity}. % ,de-cao-2021-autoregressive}.
For example, in zero-shot entity linking~\cite{logeswaran-etal-2019-zero}, textual mentions must be linked to entities unseen during training using only entity descriptions as additional data.  
Here, competitive solutions train separate BERT models to select and rank candidate entities by encoding their descriptions~\cite{logeswaran-etal-2019-zero,wu-etal-2020-scalable}.
More recently, encoder-decoder \lm{}s have been trained to retrieve entities by generating their unique names~\cite{de-cao-2021-autoregressive},  which has the advantage of scaling with the \lm{}'s vocabulary size (usually tens of thousands) instead of the \kb{} entity set size (potentially tens of millions).
\citet{de-cao-2021-autoregressive} achieve results competitive to discriminative approaches on entity linking and QA, suggesting the potential of generative entity-aware \lm{}s. 
% (see \S~\ref{entities:summary}). 
% ---albeit with higher resource requirements to train, e.g., one GPU for the classification approach proposed by~\citet{broscheit-2019-investigating}, versus 64 for the generation approach proposed by~\citet{de-cao-2021-autoregressive}. 

% In the first stage, separate \lm{}s are fine-tuned to encode textual mentions and entity descriptions for compatibility (i.e., bi-encoding, \citealp{humeau-etal-2020-poly-encoders}), such that candidate entity retrieval may be treated as a nearest-neighbor retrieval~\cite{wu-etal-2020-scalable}. 
% In the second stage, a set of candidates for each mention is retrieved, and 
% another \lm{} is fine-tuned to cross-encode concatenated mention/description sequences in order to rank the candidates~\cite{logeswaran-etal-2019-zero}.

\newpar{External entity embeddings} pretrained by a separate model have been used as strong sources of inductive bias for \lm{}s. 
For example, several variants of BERT further pretrain the base model by linearly fusing external entity embeddings with contextual word representations at the output of the BERT encoder~\cite{zhang-etal-2019-ernie,he-etal-2020-bert}. 
BERT has also been fine-tuned to match its output token representations to external entity embeddings for the task of end-to-end entity linking~\cite{chen-etal-2020-contextualized}. 
Such approaches rely heavily on the quality of the externally-learned embeddings, which is both a strength and a drawback: Such embeddings may contain  useful implicit structural information about the \kb{}, but on the other hand may propagate errors into the \lm~\cite{shen-etal-2020-exploiting}. 

% means any errors or biases in the external embeddings may be propagated and amplified in the \lm.

\subsection{Linking with middle or early fusion}
\label{entities:embeddings}

The last and strongest category of entity supervision techniques that we consider are also linking-based, but fuse entity information at earlier stages of text encoding. 
Mid-fusion approaches retrieve external entity representations in between hidden layers and re-contextualize them into the \lm{}, whereas early fusion approaches simply treat entity symbols as tokens in the vocabulary. 
Figure~\ref{fig:entity-embeddings} provides an example of input/output for early fusion. 

\newpar{Retrieving entity embeddings} and integrating them into an \lm{}'s hidden word representations is a middle-fusion technique that has the advantage of modeling flexibility: It allows the practitioner to choose where (i.e., at which layer) the entity embeddings are integrated, 
and how the entity embeddings are learned and re-contextualized into the \lm{}.
\citet{peters-etal-2019-knowledge} integrate externally pre-trained, frozen entity embeddings into BERT's final hidden layers using a word-to-entity attention mechanism.
\citet{fevry-etal-2020-entities} learn the external entity embeddings jointly during pretraining, and perform the integration in BERT's earlier hidden layers using an attention-weighted sum.
The latter approach is competitive with a 30$\times$ larger T5 \lm{} in closed-book QA (\S~\ref{entities:token-prediction}), 
suggesting that \lm{}s and \kb{}  embeddings can be trained jointly to enhance and complement each other. %  in knowledge-oriented tasks.

% perhaps be
% This may be because jointly training the embeddings with the \lm{}'s internal parameters helps the two sets of representations align closely. 

\newpar{Treating entities as ``tokens''}  by appending special reserved entity symbols to the \lm{}'s vocabulary is the earliest of entity fusion approaches  (Figure~\ref{fig:entity-embeddings}).
% This approach has the advantage of allowing the practitioner to leave the \lm{}'s internal architecture completely unchanged, if desired~\cite{rosset-etal-2020-knowledge-aware,poerner-etal-2020-e-bert}.
% However, 
For instance, 
\citet{yamada-etal-2020-luke} input entity ``tokens'' alongside textual contexts that mention these entities to RoBERTa, and use specialized word-to-entity and entity-to-entity attention matrices within its hidden layers. 
Other approaches leave the base \lm{}'s internal architecture completely unchanged and focus only on aligning the \lm{}'s word and entity embedding spaces at the input level~\cite{rosset-etal-2020-knowledge-aware,poerner-etal-2020-e-bert}. 
Note, however, that this approach may significantly enlarge the \lm{}'s vocabulary.
For example, plain BERT's vocabulary is around 30k tokens, whereas English Wikipedia has around 6 million entities.
This can make pretraining on a larger vocabulary  expensive in terms of both time and memory usage~\cite{yamada-etal-2020-luke,dufter-etal-2021-static}.

\subsection{Summary and outlook}
\label{entities:summary}

The literature on entity supervision in \lm{}s is growing rapidly. %  with little signs of letting up. 
In line with recent trends in NLP~\cite{khashabi-etal-2020-unifiedqa}, a growing number of entity supervision strategies use generative models~\cite{roberts-etal-2020-much,de-cao-2021-autoregressive}, 
which are attractive because they allow for a high level of flexibility in output and circumvent the need for classification over potentially millions of entities.
However, some studies find that generative models currently do not perform well beyond what they have memorized from the training set~\cite{wang-etal-2021-generative,lewis-etal-2021-question}.
These findings suggest that storing some entity knowledge externally (e.g., in a dense memory, \citealp{fevry-etal-2020-entities}) may be more robust, for example by allowing for efficient updates to the \lm{}'s knowledge~\cite{verga-etal-2020-facts}.
We believe that future work will need to analyze the tradeoffs  between fully-parametric and retrieval-based entity modeling in terms of pure accuracy, parameter and training efficiency, and ability to generalize beyond the training set. 

\section{Relation-level supervision}
\label{sec:relations}

Finally, we consider methods that utilize \kb{} triples or paths to supervise \lm{}s for complex, often compositional tasks like relation classification, text generation, and rule-based inference. 
We again organize methods in the order of less to more symbolic.
In this context, less symbolic approaches treat triples and paths as fully natural language  (\S~\ref{relations:templates}, ~\ref{relations:surface-forms}).
By contrast, more symbolic approaches learn distinct embeddings for relation types in the \kb{} (\S~\ref{relations:embeddings}). 
Table~\ref{table:relation} provides a taxonomy of this section with representative examples and evaluation tasks. 

% \begin{figure}[t!]
%     \centering
%     \includegraphics[width=0.99\columnwidth]{fig/relation-taxonomy.png}
%     \caption{A taxonomy of tasks and methods for relation-level supervision in \lm{}s (\S~\ref{sec:relations}), with representative examples per category. 
%     }
%     \label{fig:taxonomy-relations}
%     \vspace{-.5cm}
% \end{figure}
\begin{table*}[t!]
\centering
\caption{
    Taxonomy and representative examples of relation-level supervision in \lm{}s, with evaluation tasks conducted in the respective referenced papers. 
    \emph{Glossary of evaluation tasks}: KP---knowledge probing; ET---entity typing; RC---relation classification; QA---question answering; CR---compositional reasoning; KC---knowledge base construction; TG---text generation; GL---the GLUE family of language tasks~\cite{wang-etal-2019-glue}.
    % \reminder{add text generation}
}
\label{table:relation}
\resizebox{\textwidth}{!}{
    \begin{tabular}{ l l l c cccccccc}
    \toprule
    \multirow{2}{*}{\textbf{Relations as...}} & \multirow{2}{*}{\textbf{Supervision strategy}} & \multirow{2}{*}{\textbf{Representative examples}} && \multicolumn{8}{c}{\textbf{Evaluation task(s)}} \\ 
    \cline{5-12}
     & & && KP & ET & RC & QA & CR & KC & TG & GL \\ 
     \toprule 
     \multirow{2}{*}{Templated sentences (\S~\ref{relations:templates})} 
        & Lexicalizing triples & \cite{thorne-etal-2021-database,guan-etal-2020-knowledge-enhanced} && & &  & \cmark & \cmark & & \cmark & \\ 
        & Lexicalizing paths & \cite{clark-etal-2020-transformers,talmor-etal-2020-olmpics,talmor-etal-2020-leap} && \cmark & & & & \cmark & & & \\ 
     \midrule 
     \multirow{2}{*}{Linearized sequences (\S~\ref{relations:surface-forms})} 
        & Training on triple sequences & \cite{yao-etal-2019-kg-bert,agarwal-etal-2021-large} && \cmark & &  & \cmark & & \cmark & \cmark & \\ 
        & Injecting triples into text & \cite{liu-etal-2020-k-bert}  && & &  & \cmark & & & & \\ 
     \midrule 
     \multirow{3}{*}{Dedicated embeddings (\S~\ref{relations:embeddings})} 
        & Pooling entity representations & \cite{baldini-soares-etal-2019-matching,qin-etal-2021-erica} && & \cmark & \cmark & \cmark & & & & \\ 
        & Embedding relations externally &  \cite{wang-etal-2020-kepler,daza-etal-2021-inductive} && & \cmark & \cmark &  & & \cmark &  & \cmark \\ 
        & Treating relations as tokens & \cite{bosselut-etal-2019-comet,hwang-etal-2021-comet-atomic} && & &  & & & \cmark &  & \\ 
    \bottomrule
\end{tabular}
}
\vspace{-.3cm}
\end{table*}

\subsection{Relations as templated assertions}
\label{relations:templates}

Template-based lexicalization is a popular relation supervision strategy that does not directly expose the \lm{} to the \kb. 
Similar to how \kb{} queries are converted to cloze prompts for knowledge probing (\S~\ref{implicit:cloze}), triples are first converted to natural language assertions using relation templates, usually handcrafted.
These assertions are then fed as input to the \lm, which is trained with any number of task-specific losses.
Figure~\ref{fig:relation-sentences} provides an input/output example for this class of approach.

\newpar{Lexicalized triples} from Wikidata have been used as \lm{} training data in proof-of-concept studies demonstrating that \lm{}s can serve as natural language querying interfaces to \kb{}s under controlled conditions~\cite{heinzerling-inui-2021-language}.
A promising approach in this direction uses encoder-decoder \lm{}s to generate answer sets to natural language queries over lexicalized Wikidata triples~\cite{thorne-etal-2020-neural-databases-vldb,thorne-etal-2021-database}, toward handling multi-answer \kb{} queries with \lm{}s---thus far an understudied task in the \lm{} knowledge querying literature. 

Other approaches convert \kb{} triples to sentences using relation templates in order to construct task-specific training datasets for improved performance in, e.g.,  story generation~\cite{guan-etal-2020-knowledge-enhanced}, commonsense QA~\cite{ye-etal-2020-align-mask,ma-etal-2021-knowledge-driven}, and relation classification~\cite{bouraoui-etal-2020-inducing}.
While most of these approaches rely on template handcrafting, a few automatically mine templates using distant supervision on Wikipedia, achieving competitive results in tasks like relation classification~\cite{bouraoui-etal-2020-inducing} and commonsense QA~\cite{ye-etal-2020-align-mask}.

% representing a promising, more scalable alternative to template handcrafting. 
% Toward the task of relation classification,  \citet{bouraoui-etal-2020-inducing} extract sentences from Wikipedia containing the subject/object entities of \kb{} triples as candidate relation templates. 
% They then rank the candidates by how easily pretrained BERT can model them and fine-tune another \lm{} on the top-ranking templates to classify relations.
% \citet{ye-etal-2020-align-mask} follow a similar approach for multiple-choice commonsense QA: 
% For each entity pair in a \kb{} triple, they match the pair to Wikipedia statements containing those entities, hold out the object entity in each statement to create a question, and generate additional answer choices by sampling negative entities from the \kb. 

\newpar{Compositional paths} spanning multiple atoms of symbolic knowledge may also be lexicalized and input to an \lm{}~\cite{lauscher-etal-2020-common,talmor-etal-2020-olmpics} in order to train \lm{}s for soft compositional reasoning~\cite{clark-etal-2020-transformers,talmor-etal-2020-leap}.
Notably, when RoBERTa is fine-tuned on sentences expressing (real or synthetic) facts and rules from a \kb{}, it can answer entailment queries with high accuracy~\cite{clark-etal-2020-transformers,talmor-etal-2020-leap}.
However, as \citet{clark-etal-2020-transformers} note, these results do not necessarily confirm that \lm{}s can ``reason,'' but rather that they can at least emulate soft reasoning---raising an open question about how to develop probes and metrics to verify whether \lm{}s can actually reason compositionally. 
% Even when an \lm{} achieves high accuracy on a world knowledge benchmark requiring reasoning, 
% What kinds of probes and metrics can be developed to verify that an \lm{} is actually reasoning in a compositional manner? 
% How can we tell whether an \lm{} is actually reasoning, versus exploiting artifacts of the data or using processes otherwise unknown to us? 

% Even when an \lm{} achieves high accuracy on a world knowledge benchmark requiring reasoning, how can we tell whether an \lm{} is actually reasoning}, versus exploiting artifacts of the data or using processes otherwise unknown to us? 

% suggesting that \lm{}s may be capable of learning to ``reason'' over relational rules expressed in natural language~\cite{clark-etal-2020-transformers,talmor-etal-2020-leap}. 

\subsection{Linearizing \kb{} triples}
\label{relations:surface-forms}

\begin{figure}[t!]
    \centering
    \includegraphics[width=0.95\columnwidth]{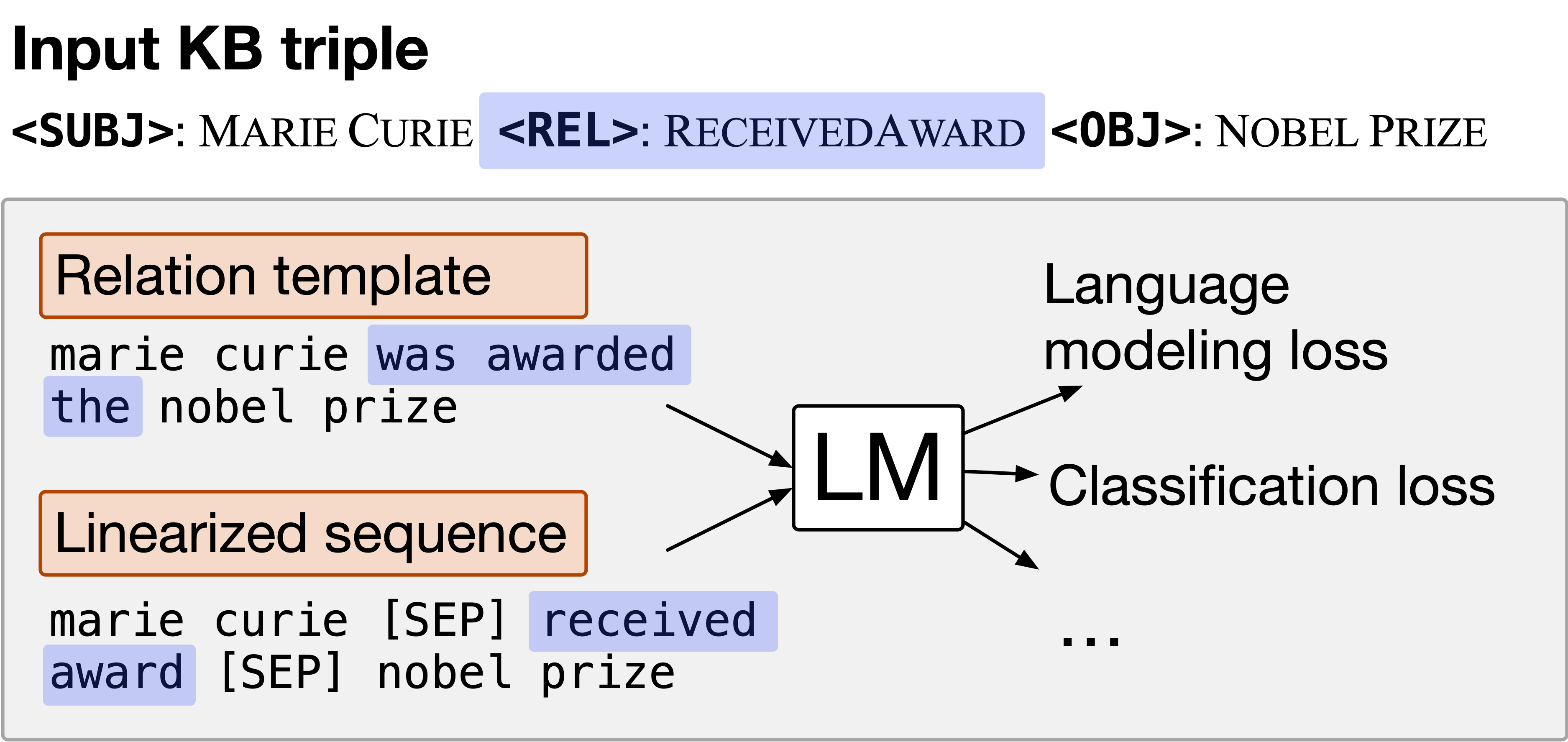}
    \caption{Strategies for representing relations as sequences: 
    Templating (\S~\ref{relations:templates}) and linearization (\S~\ref{relations:surface-forms}). 
    }
    \label{fig:relation-sentences}
    \vspace{-.5cm}
\end{figure}

The main advantage of templating is that it converts symbolic triples into sequences, which can be straightforwardly input to \lm{}s. % , since they are inherently sequence models. 
However, handcrafting templates is a manual process, and distant supervision can be noisy. % ~\cite{jiang-etal-2020-how-can}. 
To maintain the advantage of templates while avoiding the drawbacks, triples can alternatively be fed to an \lm{} by linearizing them---that is, flattening the subject, relation, and object into an input sequence (Figure~\ref{fig:relation-sentences}). 
With linearization, relation-level supervision becomes as simple as \textbf{feeding the linearized sequences} to the \lm{} and training again with task-specific losses~\cite{yao-etal-2019-kg-bert,kim-etal-2020-multi,ribeiro-etal-2020-investigating,wang-etal-2021-structure} or \textbf{injecting the sequences into the pretraining corpus}~\cite{liu-etal-2020-k-bert}.
A notable recent example of the former approach~\cite{agarwal-etal-2021-large} trains T5 on linearized Wikidata triples in order to generate  fully natural language versions of those triples. 
These verbalized triples are used as retrieval ``documents'' for improved \lm-based QA over traditional document corpora; note, however, that they can also be used as \lm{} training data for other downstream tasks in place of handcrafted templates (\S~\ref{relations:templates}). 

% \newpar{Training on triples}

% Using this approach, 
% \citet{agarwal-etal-2021-large} train to generate natural language sentences expressing Wikidata triples, then use T5 to generate a verbalized version of the entire English Wikidata.
% They demonstrate that, compared to using Wikipedia as a retrieval corpus, their verbalized version of Wikidata yields better results when combined with a retrieval-based \lm{} for question-answering. 
% Linearized \kb{} triples can also be \textbf{injected into the pretraining corpus} by inserting relation/object surface-forms adjacent to mentioned subject entities~\cite{liu-etal-2020-k-bert}. 
% In this case, specialized attention masks may be necessary to ensure that tokens from the injected triples do not attend to each other~\cite{liu-etal-2020-k-bert}.

% \citet{liu-etal-2020-k-bert} retrieve  \kb{} triples containing  entities mentioned in BERT's pretraining text, and ``stitch'' the triples into the textual mentions to create a sentence tree-like structures. 
% In the \lm{}'s attention layers, they mask select entity/token pairs to ensure that the newly-introduced ``branches'' in each sentence tree do not attend to each other. 

\begin{figure*}[t!]
    \centering
    \begin{subfigure}{.31\textwidth}
      \centering
      % include first image
      \includegraphics[width=\linewidth]{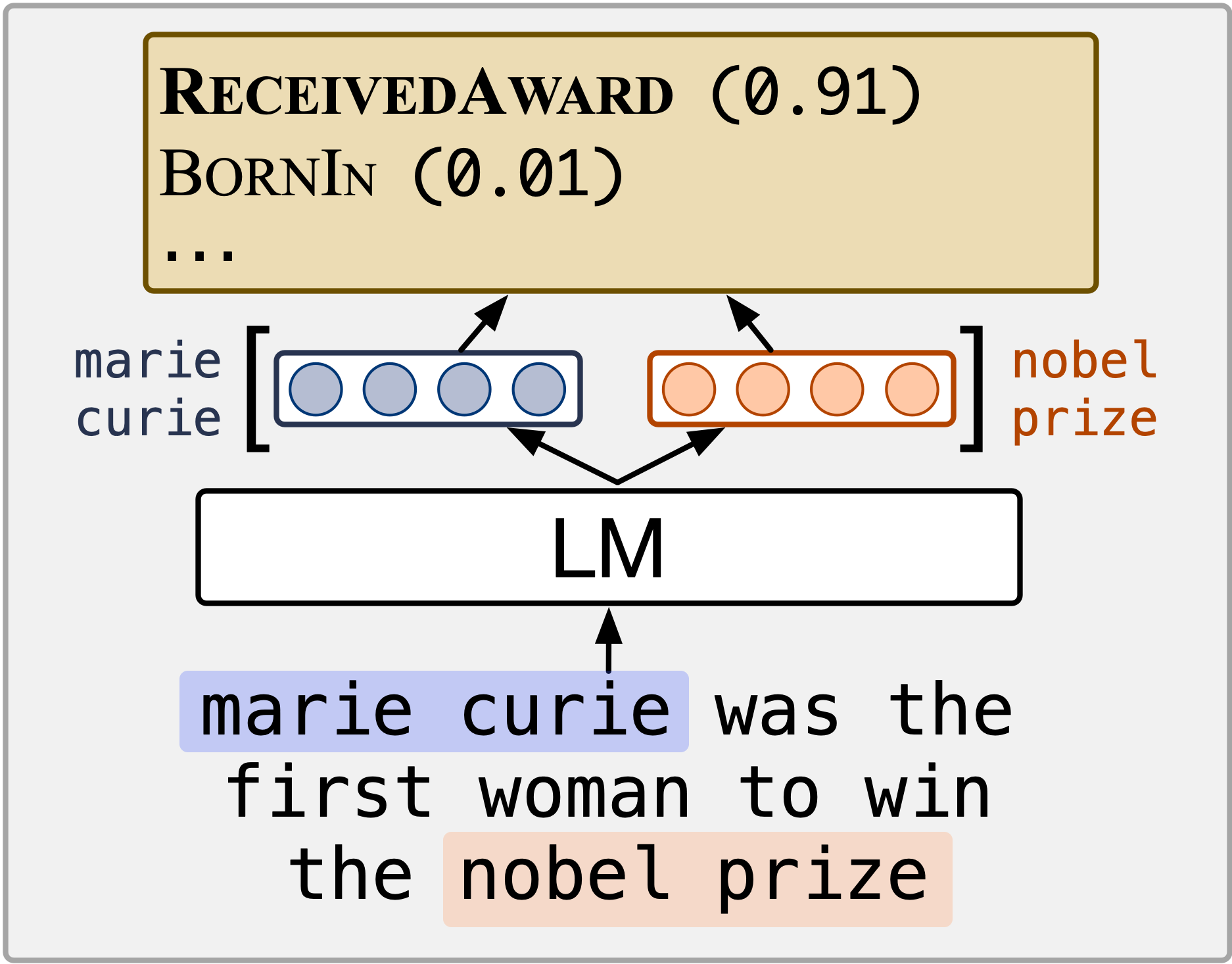}
      \caption{Late fusion---pooling}
      \label{fig:distant-supervision}
    \end{subfigure}
    ~
    \begin{subfigure}{.31\textwidth}
      \centering
      % include first image
      \includegraphics[width=\linewidth]{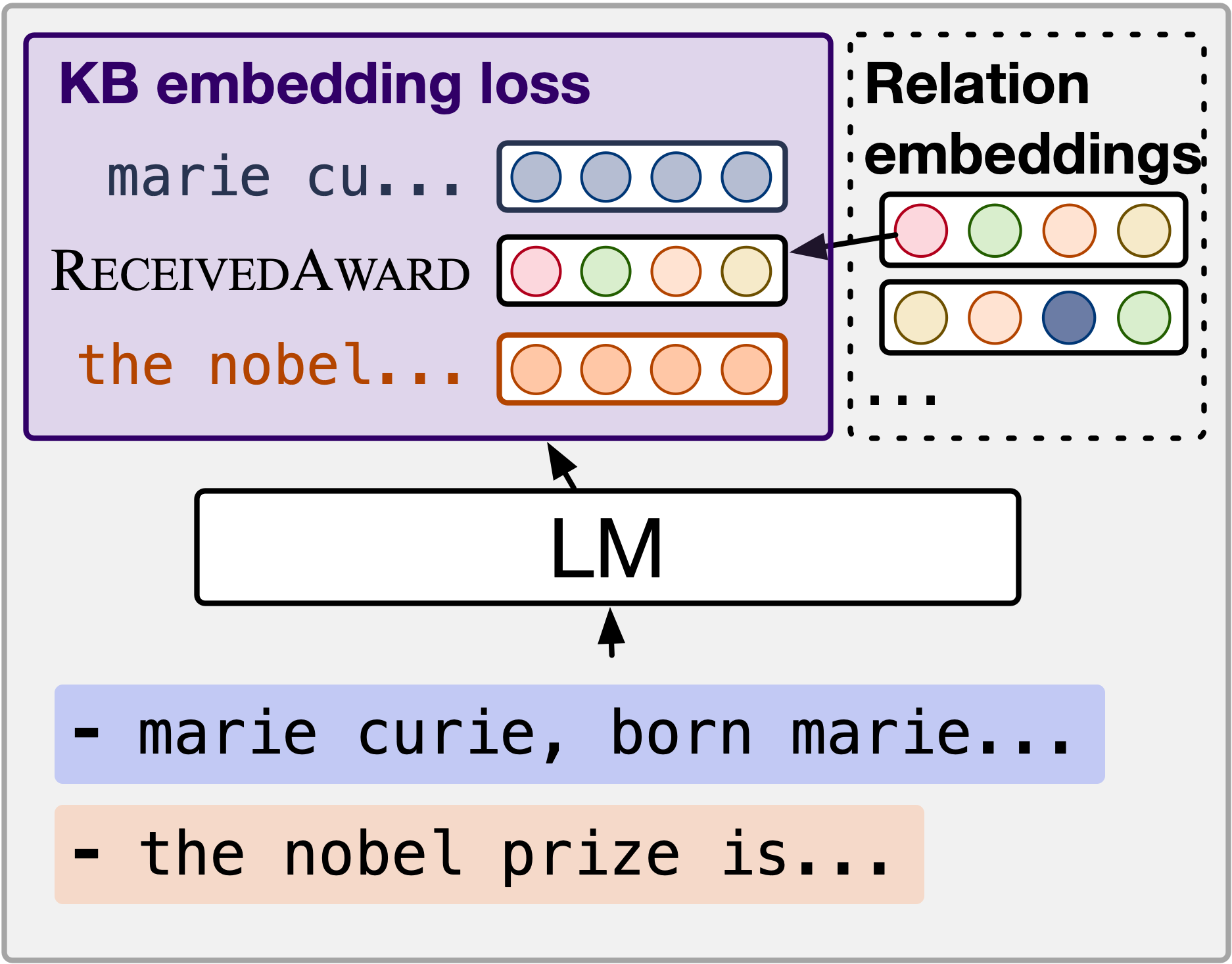}
      \caption{Late fusion---external embeddings}
      \label{fig:relation-embeddings}
    \end{subfigure}
    ~
    \begin{subfigure}{.31\textwidth}
      \centering
      % include first image
      \includegraphics[width=\linewidth]{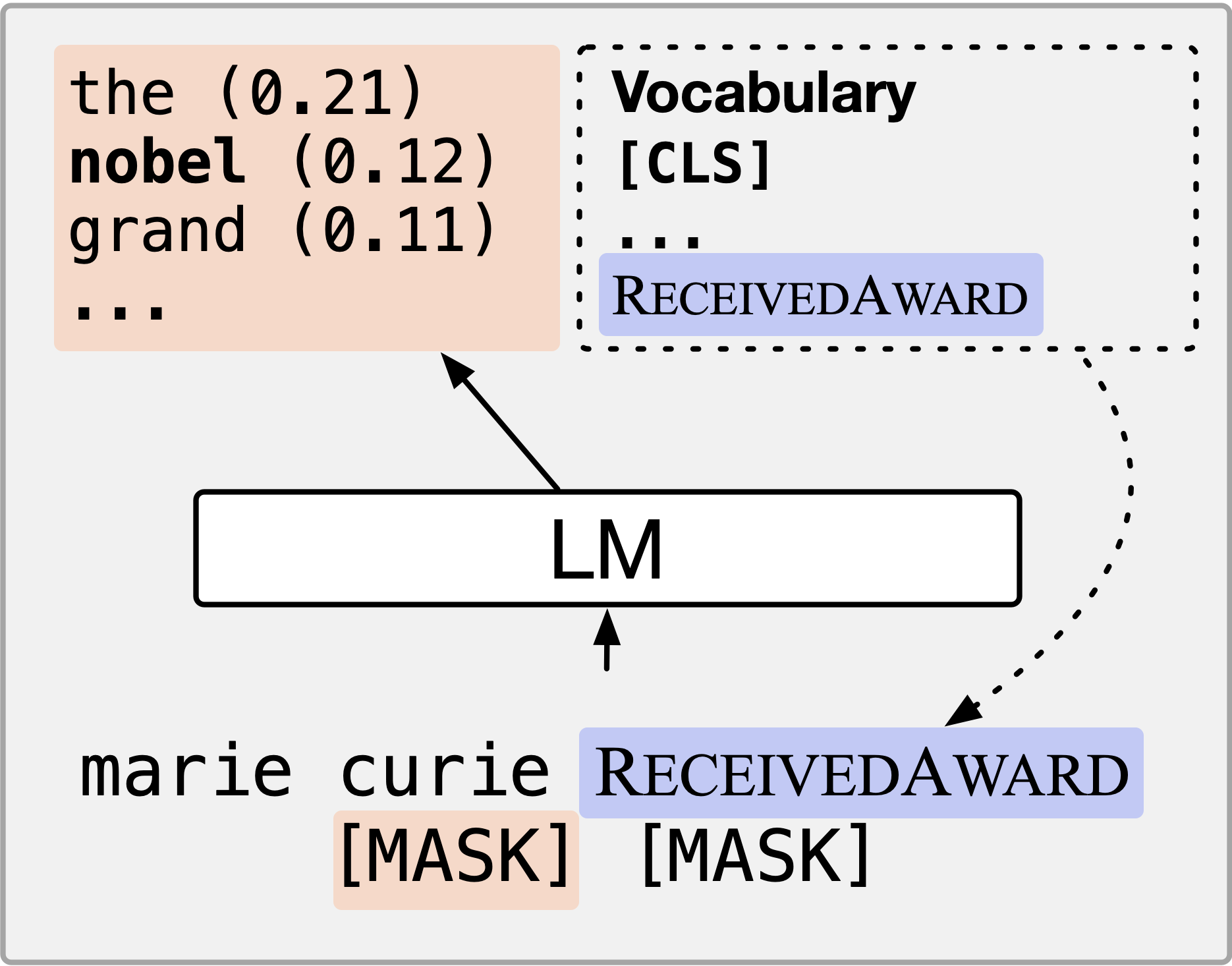}
      \caption{Early fusion---relations as ``tokens''}
      \label{fig:relations-as-tokens}
    \end{subfigure}
    \label{fig:relation-embedding-supervision}
    \caption{Examples of relation supervision strategies that incorporate dedicated embeddings of relation types. 
    }
    \vspace{-.2cm}
\end{figure*}

\subsection{Relations as dedicated embeddings}
\label{relations:embeddings}

The strategies discussed thus far treat \kb{} triples and paths as natural language sequences. 
A ``more symbolic'' approach is to represent \kb{} relation types with dedicated embeddings, and integrate these embeddings into the \lm{} using late, middle, or early fusion approaches.
Figures~\ref{fig:distant-supervision} and~\ref{fig:relation-embeddings} provide input/output examples for late fusion, whereby relation textual contexts are input to the \lm{}, and relation embeddings are constructed or integrated at the \lm{}'s output. 
Figure~\ref{fig:relations-as-tokens} exemplifies early fusion, whereby relations are treated as input tokens.

\newpar{Contextual representations of entity mention-spans may be pooled} at an \lm's output layer to represent a relation~\cite{wang-etal-2020-k-adapter,yu-etal-2020-jaket}. 
For example, \citet{baldini-soares-etal-2019-matching} concatenate the contextual representations of special entity-start markers inserted adjacent to textual entity mentions, and fine-tune BERT to output similar relation representations for statements ranging over the same entity pairs (Figure~\ref{fig:distant-supervision}). 
This approach, which proved highly successful for relation classification, has been applied to the same task in languages beyond English~\cite{koksal-ozgur-2020-relx,ananthram-etal-2020-event}, and as an additional \lm{} pretraining objective~\cite{qin-etal-2021-erica}. 

\newpar{Non-contextual relation embeddings} may be
learned by defining a separate relation embedding matrix with $|R|$ rows and fusing this matrix into the \lm{}.
One advantage of this approach, similar to methods for retrieving external entity embeddings (\S~\ref{entities:embeddings}), is that it supports fusion at both the late~\cite{wang-etal-2020-kepler,daza-etal-2021-inductive} and middle~\cite{liu-etal-2021-kg-bart} stages. 
As an example of the former, \citet{wang-etal-2020-kepler} propose an \lm{} pretraining objective whereby textual descriptions of \kb{} entities are input to and encoded by an \lm{}, then combined with externally-learned relation embeddings at the output using a link prediction loss (Figure~\ref{fig:relation-embeddings}).
Combined with standard word-level language modeling objectives, this approach enables generalization across both sentence-level tasks like relation classification, and graph-level tasks like \kb{} completion.

% at the output stage, for example with link prediction loss  , \citealp{wang-etal-2020-kepler,daza-etal-2021-inductive}), or else trained by an external model and fused into a pretrained \lm{} within \NEW{its hidden layers}~\cite{liu-etal-2021-kg-bart}.

% As an example of the former, \citet{verga-etal-2020-facts} construct a dense ``fact memory'' consisting of external entity and relation embeddings learned during \lm{} pretraining, and fuse these embeddings into the \lm{} 

% The utility of such embeddings is demonstrated by
% Toward the latter, 
% \citet{verga-etal-2020-facts} use learned entity and relation embeddings to construct a dense ``triple memory''  mapping subject/relation pairs to sets of objects. 
% This memory allows their \lm{} to retrieve and incorporate newly-injected knowledge not seen at training time, toward treating knowledge (and language) as dynamic rather than static~\cite{lazaridou-etal-2021-pitfalls}. 

\newpar{Treating relations as ``tokens,''} 
toward early fusion of relations in \lm{}s, is achieved by appending the \kb's relation types to the \lm{}'s vocabulary (Figure~\ref{fig:relations-as-tokens}). 
A notable instantiation of this approach is the  COMET commonsense \kb{} construction framework~\cite{bosselut-etal-2019-comet,hwang-etal-2021-comet-atomic,jiang-etal-2021-im}. 
Given a subject phrase/relation token as input,
COMET fine-tunes an \lm{} to generate object phrases. 
COMET demonstrates promising improvements over 400$\times$ larger \lm{}s not trained for \kb{} construction~\cite{hwang-etal-2021-comet-atomic}. 
However,  templating (\S~\ref{relations:templates}) may yield better results than adding special tokens to the vocabulary when the COMET framework is trained and tested in a few-shot setting~\cite{da-etal-2021-understanding}.

\subsection{Summary and outlook}
\label{relation:summary}

% Semantic relationships may be surfaced in many ways, which means that an \lm{} must be exposed to many realizations of a single relation in order to internalize its meaning.
% Therefore, the approach of lexicalizing relations with a small set of manually-defined templates (\S~\ref{relations:templates}), while currently popular, is likely not a scalable long-term solution to relation representation in \lm{}s. 
% We view methods that automatically convert triples to text (e.g., \citealp{agarwal-etal-2021-large}) as  more promising for scaling out complex relational inference tasks in \lm{}s. %  for example query processing~\cite{thorne-etal-2020-neural-databases-vldb} and soft reasoning~\cite{clark-etal-2020-transformers}. 

% Relation-level supervision in \lm{}s is exciting because it is a step toward machine compositional reasoning, widely thought to be key to generalization~\cite{lake-etal-2017-building}.

Relation-level supervision in \lm{}s is exciting because it enables a wide variety of complex \nlp{} tasks (Table~\ref{table:relation}). 
A unifying theme across many of these tasks is that of compositionality, or the idea that smaller ``building blocks'' of evidence can be combined to arrive at novel knowledge.
As compositionality is  thought to be key to machine generalization~\cite{lake-etal-2017-building}, we believe that further fundamental research in understanding and improving \lm{}s' soft ``reasoning'' skills (\citealp{clark-etal-2020-transformers,talmor-etal-2020-leap}, \S~\ref{relations:templates}) will be crucial. 

Finally, while most of the open directions we discuss involve improving \lm{} knowledge with \kb{}s, we find the direction of generating \kb{}s  with \lm{}s equally intriguing---reflecting the fact that \lm{}s and \kb{}s can complement each other in ``both directions,'' as automating and scaling out the construction of \kb{}s will ultimately provide \lm{}s with more relational training data. 
The generative COMET framework (\citealp{bosselut-etal-2019-comet}, \S~\ref{relations:embeddings}) has made inroads in commonsense \kb{} construction, but the same progress has not yet been observed for  encyclopedic knowledge. 
The latter entails unique challenges:
Whereas commonsense entities are not disambiguated and triples need only be plausible rather than always true, encyclopedic entities are usually disambiguated and facts are often binary true/false. 
We look forward to future research that addresses these challenges, perhaps building on recent breakthroughs in generative factual entity retrieval (\citealp{de-cao-2021-autoregressive}, \S~\ref{entities:entity-linking}).

% \section{Evaluation tasks and benchmarks}
% \label{sec:evaluation}
% \input{06-eval}

\section{Conclusion and vision}
\label{sec:conclusion}
In this review, we provide an overview of how \lm{}s may acquire relational world knowledge during pretraining and fine-tuning.
We propose a novel taxonomy that classifies knowledge representation methodologies based on the level of \kb{} supervision provided to an \lm{}, from no explicit supervision at all to entity- and relation-level supervision.
% We provide commentary and comparison between different approaches, highlighting in particular exemplary works that stand out for their accuracy and/or efficiency. 

In the future, we envision a stronger synergy between the perspectives and tools from the language modeling and knowledge bases communities. %  as each brings a well-developed set of perspectives and tools that may complement the other.
In particular, we expect powerful and expressive \lm{}s, which are actively being developed in \nlp{}, to be increasingly combined with large-scale \kb{} resources to improve their knowledge recall and reasoning abilities. 
On the converse, we expect such \kb{} resources to be increasingly generated directly by \lm{}s. 
Within both of these directions, we hope that future work will continue to explore the themes discussed in this paper, in particular that of delineating and testing \kb-level memorization versus generalization in \lm{}s.
% better characterize the accuracy and efficiency tradeoffs of storing relational knowledge fully internally within an \lm{} (i.e., within its parameters) versus external to the \lm{} (i.e., as a retrieval corpus). 
We also expect that more standardized benchmarks and tasks for evaluating \lm{} knowledge will be developed, a direction that has recently seen some progress~\cite{petroni-etal-2021-kilt}.
As research at the intersection of \lm{}s and \kb{}s is rapidly progressing, we look forward to new research that better develops and combines the strengths of both knowledge representations.

\section*{Acknowledgements}
We thank the reviewers for their thoughtful feedback.
This material is based upon work supported by the National Science Foundation under a Graduate Research Fellowship and CAREER Grant No.~IIS 1845491, the Advanced Machine Learning Collaborative Grant from Procter \& Gamble, and an Amazon faculty award. 
% Any opinions, findings, and conclusions or recommendations expressed in this material are those of the author(s) and do not necessarily reflect the views of the National Science Foundation or other funding parties.

\balance
\bibliography{references/a-f,references/g-l,references/m-q,references/r-v,references/w-z}
\bibliographystyle{acl_natbib}

\end{document}